\documentclass{article} 
\usepackage{iclr2027_conference,times}

\usepackage[utf8]{inputenc} 
\usepackage[T1]{fontenc}    
\usepackage{hyperref}       
\usepackage{url}            
\usepackage{booktabs}       
\usepackage{amsfonts}       
\usepackage{amsmath}        
\usepackage{amssymb}        
\usepackage{graphicx}       
\usepackage{wrapfig}        
\usepackage{subcaption}     
\usepackage{float}          
\usepackage{pifont}         
\usepackage{nicefrac}       
\usepackage{microtype}      
\usepackage{xcolor}         
\usepackage{colortbl}       
\definecolor{opsgreen}{HTML}{2B9E3F}
\definecolor{opsred}{HTML}{C0392B}
\newcommand{\cmark}{\textcolor{opsgreen}{\ding{52}}}
\newcommand{\xmark}{\textcolor{opsred}{\ding{56}}}
\usepackage{listings}       
\lstdefinestyle{promptbox}{
  basicstyle=\ttfamily\footnotesize,
  breaklines=true,
  breakatwhitespace=false,
  breakautoindent=false,
  breakindent=0pt,
  columns=fullflexible,
  frame=single,
  framerule=0.4pt,
  xleftmargin=4pt,
  xrightmargin=4pt,
  aboveskip=4pt,
  belowskip=4pt,
  upquote=true
}
\newcommand{\promptlabel}[1]{\medskip\noindent\textbf{#1}\par\nobreak\vspace{2pt}}

\title{Hindsight Compacts but Does Not Repair:\\Rethinking On-Policy Self-Distillation\\in Reasoning Models}

\author{Jaehoon Kim \& Dongha Lee\thanks{Corresponding author.} \\
Yonsei University \\
\texttt{\{jaeh8nkim,donalee\}@yonsei.ac.kr}
}

\iclrfinalcopy 
\begin{document}

\maketitle
\lhead{Preprint.}

\begin{abstract}
  On-Policy Self-Distillation (OPSD) has emerged as a promising post-training method: requiring only privileged context, it enables token-level credit assignment from a self-teacher with hindsight, yielding higher accuracy with shorter responses.
However, in thinking-enabled mathematical reasoning, where solving a problem takes long reasoning traces that explore, reflect, and backtrack, its length reductions persist but its accuracy gains largely do not.
Does OPSD's hindsight signal, then, still repair failed trajectories, or does it only compact viable ones?
We hypothesize compaction: in long traces, hindsight reveals which steps were unnecessary more readily than which steps would have repaired the solution.
To test this, we apply OPSD separately to correct and incorrect rollouts.
Training only on correct rollouts shortens responses by 18--29\% while largely preserving accuracy, whereas training only on incorrect rollouts degrades it.
The accuracy gap holds across three models, six benchmarks, and three seeds, and under a same-prompt control for difficulty.
Neither branch raises the pass@$k$ ceiling, and both suppress exploration and reflection markers, which viable traces can spare but failed ones need.
Changing the divergence, enriching or reinjecting the privileged context, and training longer only move OPSD along the same accuracy-length tradeoff.
Hindsight compacts reasoning the model can already produce but does not repair reasoning it cannot.

\end{abstract}

\begin{figure}[H]
  \makebox[\linewidth][c]{%
    \begin{minipage}{1.0\linewidth}
      \centering
      \includegraphics[width=1.0\linewidth]{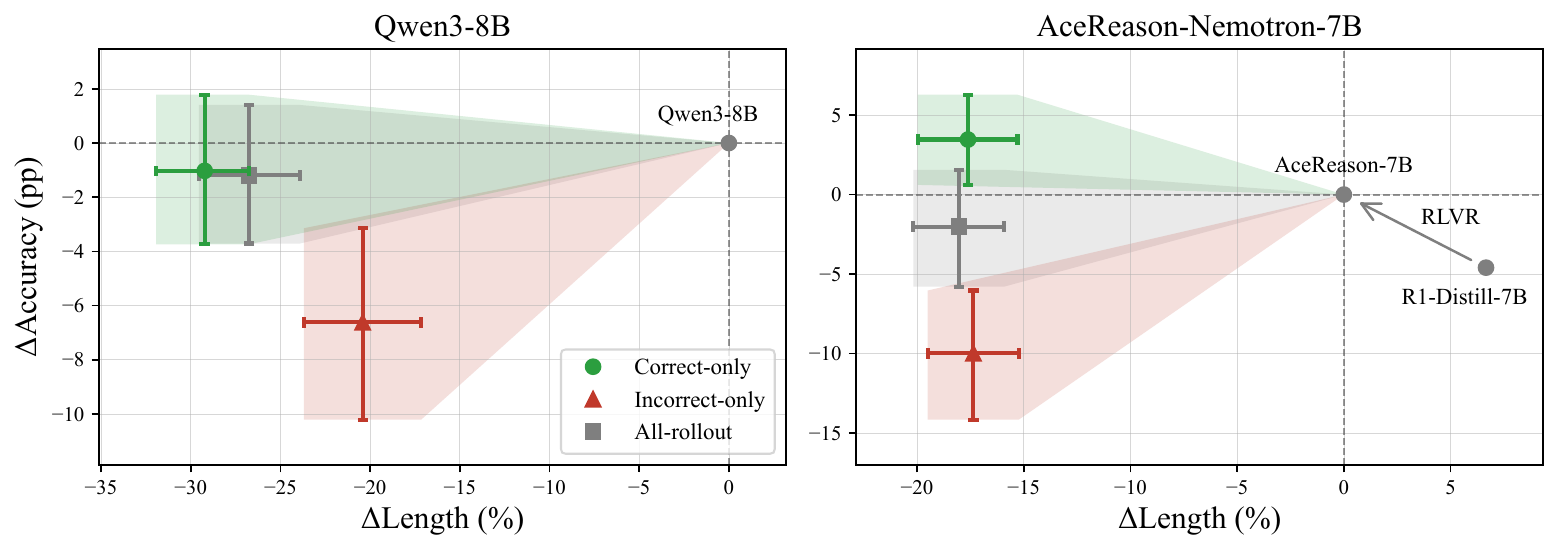}
      \caption{Impact of OPSD in the accuracy-length plane. Each point reports average $\Delta$Accuracy and $\Delta$Length over MATH500, AIME24, and AIME25 relative to each model's pre-OPSD baseline; the horizontal and vertical bars at each point form a cross showing 95\% paired bootstrap confidence intervals. \emph{Correct-only} and \emph{Incorrect-only} restrict OPSD training to correct or to incorrect rollouts, while \emph{All-rollout} trains on every rollout. For reference, the arrow in the right panel marks the RLVR step from R1-Distill-7B to AceReason-7B. \textbf{Takeaway:} on both models, \emph{Correct-only} largely preserves accuracy while substantially shortening responses, whereas \emph{Incorrect-only} degrades accuracy without shortening them more. This contrast indicates that OPSD compacts rather than repairs.}
      \label{fig_acc_len_tradeoff}
    \end{minipage}
  }
\end{figure}

\section{Introduction}\label{sec_intro}

On-Policy Self-Distillation (OPSD) has recently emerged as a promising post-training method \citep{DBLP:journals/corr/abs-2601-18734,DBLP:journals/corr/abs-2601-20802,DBLP:journals/corr/abs-2602-12275,arxiv_2603_05433}.
Its appeal is that it needs no external teacher, only privileged context: a self-teacher that shares the student's parameters is conditioned on information the student does not see, such as an expert demonstration, environment feedback, or the student's own reflection.
For each token in a student-generated rollout, the self-teacher provides a target distribution, which the student is trained to match.
The result is token-level credit assignment, in contrast to Reinforcement Learning with Verifiable Rewards (RLVR), which scores an entire rollout with a scalar outcome reward and applies the resulting advantage uniformly across its tokens \citep{DBLP:journals/corr/abs-2402-03300,DBLP:journals/corr/abs-2501-12948,DBLP:journals/corr/abs-2503-14476}.
OPSD-trained models are accordingly reported to reach higher accuracy with shorter responses.

In thinking-enabled mathematical reasoning, however, where solving a problem takes long reasoning traces that explore, reflect, and backtrack, the accuracy gains largely do not carry over.
OPSD-family methods report large gains with thinking disabled but small or negative gains with thinking enabled, while length reductions persist (Table~\ref{tab_method_comparison}).
This disparity suggests a teacher who knows, in hindsight, how the problem should have gone can act on a rollout in two ways.
It could repair a failed trajectory, marking where the reasoning went wrong and what should have come instead.
Or it could compact a viable trajectory, marking which steps were unnecessary.
We therefore ask whether, in long traces, the hindsight-guided signal still repairs a failed trajectory or only compacts a viable one.

We hypothesize compaction: by specifying what the answer should be, privileged context gives an informative learning signal on short outputs but not on long traces.
A thinking-disabled output is short and answer-shaped, so a teacher that knows the answer can specify the desired tokens nearly one for one.
A thinking-enabled trace instead explores, reflects, and backtracks along a path whose desired tokens are hard for the teacher to specify \citep{DBLP:journals/corr/abs-2201-11903,DBLP:journals/corr/abs-2211-14275,DBLP:journals/corr/abs-2305-20050}.
In such a trace, \textbf{hindsight reveals which steps were unnecessary more readily than which steps would have repaired the solution}.
On a viable trajectory, the signal trims redundancy.
On a failed one, it has nothing to repair with: it suppresses the reconsiderations viable traces can spare but failed ones need \citep{DBLP:journals/corr/abs-2603-24472}, making the path more confident rather than more correct.

To test this, we apply OPSD separately to correct and incorrect rollouts, holding the privileged context and loss fixed.
Correctness decides only which rollouts receive the loss, not which tokens are targeted, meaning neither branch sees an outcome reward.
If hindsight repairs, training on incorrect rollouts, where repair is needed, should raise accuracy.
If it only compacts, training on correct rollouts should shorten responses safely, while training on incorrect ones should hurt accuracy.

Our results show that \textbf{in thinking-enabled mathematical reasoning, hindsight compacts but does not repair}.
\emph{Correct-only} OPSD shortens responses by 29\% on Qwen3-8B at a 1.0-point accuracy cost and by 18\% on AceReason-Nemotron-7B with a 3.5-point gain, while \emph{Incorrect-only} loses 6.6 and 10.0 points (Figure~\ref{fig_acc_len_tradeoff}).
The accuracy gap holds across three models, six benchmarks, and three seeds, and under a same-prompt control for difficulty.
Neither branch raises the pass@$k$ ceiling, so no new solutions emerge.
Both branches suppress exploration and reflection markers such as ``wait'' and ``maybe'', the reconsiderations that viable traces can spare but failed ones need.

We then ask whether narrower factors hide a repair effect.
The reverse-KL direction may fail to pull a failed trace toward the teacher's alternatives; the teacher may lack enough privileged information; its signal may fade before the mid-trace steps where failed trajectories need to change, since the privileged context appears only at the start of the teacher's input; or repair may need longer training to accumulate.
Switching to symmetric or split-direction divergences, enriching the privileged context, reinjecting it mid-trace, and training five times longer each move OPSD along the same accuracy-length tradeoff, but none makes it repair failed trajectories.

These findings call for rethinking what OPSD contributes to reasoning models.
Its accuracy and length effects should be evaluated separately: repairing failed reasoning requires a signal that can supply reasoning states the model does not yet produce, whereas compacting viable reasoning only requires pruning those it already produces.
Compaction is also cheap: OPSD reaches most of its compression early in training, and at a matched rollout budget it compresses about four times as much as length-penalized GRPO.
OPSD is thus better understood as a way to compact reasoning than as a way to repair it.

\section{Background and Related Work}\label{sec_related}

\subsection{On-Policy Self-Distillation}\label{sec_opsd}

On-Policy Self-Distillation trains a student against soft targets from a teacher \citep{DBLP:journals/corr/HintonVD15,DBLP:conf/iclr/AgarwalVZSGGB24} that shares the student's parameters and is conditioned on extra context not given to the student \citep{DBLP:journals/corr/abs-2112-00861}.
For a prompt $x$, the student sees only $x$ and produces a rollout $y\sim\pi_S(\cdot\mid x)$; the teacher sees $x$ plus a privileged context $c$ (a ground-truth answer, an expert demonstration, or a written reflection) and scores the student at each prefix $y_{<t}$ via reverse-KL:
\begin{equation}\label{eq_opsd}
\mathcal{L}_{\mathrm{OPSD}}
=
\mathbb{E}_{x,\;y\sim \pi_S(\cdot\mid x)}
\left[
\frac{1}{|y|}
\sum_{t=1}^{|y|}
D_{\mathrm{KL}}
\left(
\pi_S(\cdot\mid x,y_{<t})
\;\|\;
\pi_T(\cdot\mid x,c,y_{<t})
\right)
\right].
\end{equation}
In Equation~\ref{eq_opsd}, the student policy determines the state distribution through $y\sim\pi_S(\cdot\mid x)$, while the privileged teacher determines the target distribution through $\pi_T(\cdot\mid x,c,y_{<t})$; the objective requires privileged context for the teacher but no rollout correctness labels.
Reverse-KL averages over the student's support and acts on tokens the student already assigns mass to \citep{arxiv_2603_05433}.

\subsection{Related Work}\label{sec_prior_work}

\paragraph{OPSD as a post-training mechanism.}
A growing body of work uses OPSD as a post-training mechanism, either as the main training objective \citep{DBLP:journals/corr/abs-2601-18734,arxiv_2603_05433,DBLP:journals/corr/abs-2602-12275,DBLP:journals/corr/abs-2601-20802,DBLP:journals/corr/abs-2601-19897} or as a supplementary loss layered on top of another optimizer such as GRPO \citep{DBLP:journals/corr/abs-2602-04942,DBLP:journals/corr/abs-2602-20574,DBLP:journals/corr/abs-2603-10165,DBLP:journals/corr/abs-2603-23871,arxiv_2604_02288,arxiv_2604_03128,arxiv_2604_12002}.
However, most reported gains are measured with thinking turned off, on instruct models or on reasoning models with thinking disabled, even on mathematical benchmarks.
This distinction is consequential: a thinking-disabled output is short and answer-shaped, so privileged context specifies it nearly token for token and per-token KL installs that mapping directly; a thinking-enabled trace is long and free-form, exploring, reflecting, and backtracking along a path the answer does not determine, so a hindsight-conditioned self-teacher cannot pinpoint which tokens would have improved it.
Consistent with this, OPSD-family methods report large gains with thinking disabled but small or negative gains with thinking enabled on mathematical reasoning (Table~\ref{tab_method_comparison}).
The most direct account of this shortfall attributes it to exploration suppression: a self-teacher conditioned on richer context verbalizes less uncertainty, and distilling it strips the student of the checks and reconsiderations that long-horizon reasoning relies on \citep{DBLP:journals/corr/abs-2603-24472}.
That account, which we adopt, explains why self-distillation can degrade reasoning but not whether the same hindsight-guided signal ever repairs a failed trajectory or only compacts a viable one.
Our work answers this question by applying OPSD separately to correct and incorrect rollouts, which separates the two effects.

\begin{table}[H]
  \caption{Accuracy and length deltas for representative OPSD-family methods on 7--8B-class models. Numbers are each paper's own headline measurements on its chosen benchmarks; $\dagger$ indicates our measurements averaged over MATH500, AIME24, and AIME25.}
  \label{tab_method_comparison}
  \vspace{3pt}
  \centering
  \footnotesize
  \begin{tabular}{lcccc}
    \toprule
    \textbf{Method} & \textbf{Thinking-enabled} & \textbf{Math-reasoning} & {\boldmath\textbf{$\Delta$Acc. (pp)}} & {\boldmath\textbf{$\Delta$Len. (\%)}} \\
    \midrule
    SDFT \citep{DBLP:journals/corr/abs-2601-19897}        & \xmark & \xmark & $+25$  & not reported \\
    SDPO \citep{DBLP:journals/corr/abs-2601-20802}        & \xmark & \xmark & $+24$  & $-68.8$ \\
    OPSD \citep{DBLP:journals/corr/abs-2601-18734}        & \cmark & \cmark & $+2.2$ & not reported \\
    CRISP$^{\dagger}$ \citep{arxiv_2603_05433}            & \cmark & \cmark & $-2.5$ & $-22.8$ \\
    \emph{Correct-only} OPSD (ours)                     & \cmark & \cmark & $-1.0$ & $-29.2$ \\
    \bottomrule
  \end{tabular}
\end{table}

\paragraph{Reasoning trace compaction.}
A parallel line of work treats long reasoning traces themselves as the object to compress.
One approach builds shorter correct traces by rewriting, selecting, pruning, or self-sampling compressed chains of thought and fine-tunes on them \citep{arxiv_2509_14093,DBLP:conf/emnlp/XiaLWLL25,arxiv_2602_01982}.
Another approach modifies RL training with length penalties so that correctness is rewarded jointly with shorter reasoning \citep{arxiv_2503_04697,arxiv_2504_01296,arxiv_2601_21418}.
Our setting is different: OPSD has no explicit length objective or outcome reward, and we compare it against both families at matched rollout budget in Section~\ref{sec_longer_training}.

\section{Compaction of Viable Rollouts, Not Repair of Failed Ones}\label{sec_correctness}

\subsection{Experimental Setup}\label{sec_setup}

Our experiments isolate where the OPSD signal is applied: in the main comparison, we keep the privileged context fixed and vary which rollouts contribute to training or which divergence is used.

\paragraph{Privileged context.}
Our default privileged context is \emph{Reflection from GT}, a student-generated reflection on the rollout produced after seeing the ground-truth answer.
To construct it, the student attempts the problem, then receives the ground-truth answer and reflects on its attempt; the self-teacher is conditioned on the problem and this reflection.
The privileged context is held fixed throughout this section and varied in Section~\ref{sec_ablations}.

\paragraph{Models and data.}
We evaluate OPSD on three 7--8B thinking-enabled reasoning models: Qwen3-8B \citep{DBLP:journals/corr/abs-2505-09388}, AceReason-Nemotron-7B \citep[AceReason-7B;][]{DBLP:journals/corr/abs-2505-16400}, a public checkpoint trained with GRPO \citep{DBLP:journals/corr/abs-2402-03300} on top of DeepSeek-R1-Distill-Qwen-7B \citep[R1-Distill-7B;][]{DBLP:journals/corr/abs-2501-12948}, and R1-Distill-7B itself, the pre-RLVR base of AceReason-7B, which adds a model without RL training.
We report Qwen3-8B and AceReason-7B in the main text and R1-Distill-7B in Appendix~\ref{app_lineage}.
All OPSD runs use DAPO-Math-17K \citep{DBLP:journals/corr/abs-2503-14476} as the on-policy training prompt distribution, and the paper's claims are scoped throughout to thinking-enabled mathematical reasoning on these models and this distribution.

\paragraph{Evaluation.}
We evaluate on MATH500 \citep{DBLP:journals/corr/abs-2103-03874}, AIME24, and AIME25, which together test mathematical reasoning across a moderate-to-hard difficulty range, and extend to AIME26, HMMT26, and OlympiadBench in Appendix~\ref{app_benchmarks}.
Each benchmark is scored using average@8 accuracy, sampled at temperature 0.6, and mean response length under a 30{,}000-token generation cap.
We report $\Delta$Accuracy in percentage points and $\Delta$Length as a relative percentage change from each model's own pre-OPSD baseline.

\paragraph{Training protocol.}
Unless otherwise noted, all runs use AdamW with batch size 32, maximum training generation length 8{,}192, constant learning rate $1\times10^{-6}$, gradient clipping at 1.0, no warmup, and 100 optimization steps.
The self-teacher is held frozen at the student's initial weights throughout training, following prior work \citep{DBLP:journals/corr/abs-2601-18734,DBLP:journals/corr/abs-2603-24472}.
Student rollouts during training are sampled at temperature 0.7.
A rule-based verifier labels each rollout correct or incorrect; the label is used only by the outcome-filtered settings below, and \emph{All-rollout} OPSD does not require it.

\subsection{Outcome-Filtered OPSD Training Reveals Compaction Rather Than Repair}\label{sec_results}

Correctness enters OPSD only as a diagnostic: standard \emph{All-rollout} OPSD needs no correctness labels, only privileged context for the self-teacher.
We introduce outcome-filtered training as the intervention: \emph{Correct-only} and \emph{Incorrect-only} use the same reverse-KL loss on correct and incorrect rollouts, respectively.
If OPSD repairs failed trajectories, \emph{Incorrect-only} should improve accuracy, since those are precisely the cases where repair is needed.
If OPSD mainly compacts viable trajectories, \emph{Correct-only} should be the safest compaction setting.

Accuracy depends strongly on which rollouts receive KL; compression far less so.
On Qwen3-8B, \emph{All-rollout} and \emph{Correct-only} both shorten responses, but \emph{Correct-only} compacts them more safely, while \emph{Incorrect-only} loses 6.6 pp on average and compresses the least.
On AceReason-7B, the pattern sharpens: the average accuracy delta moves from $+3.5$ pp under \emph{Correct-only} to $-10.0$ pp under \emph{Incorrect-only} with the same privileged context (Table~\ref{tab_main}).

Rollout correctness is entangled with prompt difficulty: correct rollouts come disproportionately from easier prompts and incorrect ones from harder prompts, so the two branches also train on different prompt distributions.
To separate the two, we train the branches in lockstep on the same prompts, admitting a prompt at each step only if the \emph{Correct-only} model produces at least one correct rollout and the \emph{Incorrect-only} model at least one incorrect rollout within four attempts (Appendix~\ref{app_lockstep}).
Matching prompts preserves the ordering and most of the gap: on Qwen3-8B, \emph{Correct-only} moves to $+0.4$ pp at $-23.3\%$ length and \emph{Incorrect-only} to $-3.4$ pp at $-17.9\%$, a 3.9 pp accuracy gap against 5.6 pp unmatched.
Difficulty explains part of the original gap; the rest is the viability of the rollouts the KL objective lands on.

\begin{table}[H]
  \caption{Outcome-filtered OPSD training and divergence comparison on Qwen3-8B and AceReason-Nemotron-7B; \emph{Avg.\ $\Delta$} averages per-benchmark changes from baseline. \emph{Correct-only} compacts most safely, \emph{Incorrect-only} damages accuracy, and divergence variants do not restore robust repair.}
  \label{tab_main}
  \vspace{3pt}
  \centering
  \footnotesize
  \setlength{\tabcolsep}{2pt}
  \begin{tabular}{lcccccccc}
    \toprule
    & \multicolumn{4}{c}{\textbf{Accuracy}} & \multicolumn{4}{c}{\textbf{Length}} \\
    \cmidrule(lr){2-5} \cmidrule(lr){6-9}
    \textbf{Method} & \textbf{MATH500} & \textbf{AIME24} & \textbf{AIME25} & {\boldmath\textbf{Avg.\ $\Delta$\ (pp)}} & \textbf{MATH500} & \textbf{AIME24} & \textbf{AIME25} & {\boldmath\textbf{Avg.\ $\Delta$\ (\%)}} \\
    \midrule
    \rowcolor{gray!15}\multicolumn{9}{l}{\textbf{Qwen3-8B}} \\
    Baseline           & $78.1$ & $70.8$ & $62.9$ & ---     & $4{,}849$ & $14{,}342$ & $17{,}031$ & ---      \\
    All-rollout     & $77.9$ & $69.6$ & $60.8$ & $-1.19$ & $3{,}419$ & $10{,}878$ & $12{,}499$ & $-26.75$ \\
    Correct-only    & $78.4$ & $72.1$ & $58.3$ & $-1.03$ & $3{,}208$ & $10{,}365$ & $12{,}587$ & $-29.22$ \\
    Incorrect-only  & $74.1$ & $68.3$ & $49.6$ & $-6.62$ & $3{,}957$ & $10{,}986$ & $13{,}723$ & $-20.41$ \\
    Split-direction & $76.8$ & $75.8$ & $60.8$ & $+0.54$ & $3{,}570$ & $10{,}419$ & $12{,}983$ & $-25.83$ \\
    All-rollout JSD    & $77.2$ & $71.7$ & $58.8$ & $-1.42$ & $3{,}416$ & $10{,}835$ & $12{,}531$ & $-26.81$ \\
    \midrule
    \rowcolor{gray!15}\multicolumn{9}{l}{\textbf{AceReason-Nemotron-7B}} \\
    Baseline           & $68.5$ & $43.3$ & $32.5$ & ---      & $3{,}206$ & $9{,}014$ & $10{,}104$ & ---      \\
    All-rollout     & $67.1$ & $38.3$ & $32.9$ & $-2.02$  & $2{,}505$ & $7{,}769$ & $8{,}241$  & $-18.03$ \\
    Correct-only    & $71.0$ & $46.2$ & $37.5$ & $+3.46$  & $2{,}420$ & $7{,}880$ & $8{,}515$  & $-17.61$ \\
    Incorrect-only  & $54.0$ & $33.3$ & $27.1$ & $-10.00$ & $2{,}625$ & $7{,}714$ & $8{,}129$  & $-17.35$ \\
    Split-direction & $68.2$ & $42.1$ & $34.2$ & $+0.04$  & $2{,}524$ & $7{,}337$ & $7{,}835$  & $-20.77$ \\
    All-rollout JSD    & $68.7$ & $44.2$ & $38.8$ & $+2.42$  & $2{,}443$ & $7{,}478$ & $8{,}084$  & $-20.27$ \\
    \bottomrule
  \end{tabular}
\end{table}

If OPSD repaired failed reasoning, it should raise the capability ceiling: high-$k$ pass@$k$ measures whether a correct solution is reachable within $k$ attempts, and repair adds solves whereas compaction does not.
Averaged over six benchmarks (Appendix~\ref{app_passk}), no branch raises pass@8 on either model.
\emph{Incorrect-only} sits below the base at every $k$, and \emph{Correct-only} tracks the base on Qwen3-8B ($-0.6$ at pass@8) while on AceReason-7B it raises pass@1 by 3.2 pp and lowers pass@8 by 4.0 pp, the signature of concentrating probability mass on already-viable paths rather than adding new ones.

At the question level, \emph{Incorrect-only} moves specific questions toward incorrectness, and training on all rollouts yields no net repair on the evaluation set (Table~\ref{tab_correctness_shift}).
The same branch ordering appears on AIME26, HMMT26, and OlympiadBench for both models (Appendix~\ref{app_benchmarks}), on R1-Distill-7B (Appendix~\ref{app_lineage}), and across additional training seeds (Appendix~\ref{app_seeds}).

The branch ordering raises a natural question: if \emph{Correct-only} OPSD trains only on correct rollouts, is it just rejection-sampling fine-tuning?

\subsection{Filtering Correct Rollouts Does Not Mean Reinforcing Correct Tokens}\label{sec_filtering}

The outcome-filtered training result may make \emph{Correct-only} OPSD look like rejection-sampling fine-tuning (RFT) \citep{DBLP:journals/corr/abs-2203-14465,DBLP:journals/corr/abs-2312-06585}, but the objectives differ in where correctness enters the update.
In outcome-filtered OPSD, rollout correctness changes which prefixes enter the KL loss, not which token is reinforced at a state: correctness gates states, not token targets.

Under RFT the target is the sampled correct token, $\sum_t \log \pi_S(a_t^{\mathrm{correct}}\mid s_t)$, with rejection sampling choosing which state-target pairs the loss sees; GRPO with positive advantage likewise pushes up the chosen token, scaling $\nabla \log \pi_S(a_t\mid s_t)$ by $A(\tau)>0$ over the whole rollout.
OPSD takes the same local form, but its per-token advantage is the teacher-student log-probability gap rather than the rollout reward:
\begin{equation}\label{eq_policy_update}
\nabla_\theta \mathcal{J}(\theta)
=
\mathbb{E}_{x,\;y\sim\pi_S(\cdot\mid x)}
\left[
\sum_{t=1}^{|y|} \widehat{A}_t
\nabla_\theta \log \pi_S(y_t\mid s_t)
\right],
\end{equation}
with the per-token advantage $\widehat{A}_t$ instantiated per method,
\begin{equation}\label{eq_token_advantage}
\begin{aligned}
\widehat{A}^{\mathrm{GRPO}}_t = A(\tau),
\qquad
\widehat{A}^{\mathrm{OPSD}}_t
=
\log \pi_T(y_t\mid s_t,c)
-
\log \pi_S(y_t\mid s_t).
\end{aligned}
\end{equation}

The OPSD term is an implicit token reward, positive when the privileged-context teacher assigns $y_t$ more probability than the student: a teacher-preference reward, not an outcome reward, so a token can be encouraged for matching the teacher even on a failed trajectory.

Outcome filtering therefore gives OPSD no channel for reinforcing the tokens that produced a correct answer; its role is to diagnose where teacher-following compaction is safe.
Empirically, RFT on the shorter half of verified-correct rollouts shortens responses by only 2.6\% (Section~\ref{sec_longer_training}).

\subsection{Correctness Determines Whether Compaction Is Safe}\label{sec_safe_compression}

If correctness does not choose the token target, where does it matter?
It determines the substrate on which the same hindsight-guided and confident self-teacher signal lands.
On correct rollouts, it can remove redundant checks, over-verification, and self-sabotaging detours while preserving the solution structure.
One plausible reading is the compounding-error account \citep{arxiv_2603_05433}: when per-token reasoning is approximately reliable, redundant tokens still carry small accumulating risk, so removing them can improve reliability without changing per-token competence.

The same hindsight-guided signal is less safe on incorrect rollouts.
Privileged context can concentrate the teacher-student gap at pivot tokens, where knowing the answer makes one path look obvious even though the unconditioned student may still need to explore \citep{arxiv_2604_03128}.
When the rollout is wrong, those pivots are often precisely where exploration, checking, or recovery moves are needed.
Compressing them can therefore make the failed path more confident and less exploratory, suppressing the reconsiderations that might have helped the student escape \citep{DBLP:journals/corr/abs-2603-24472}.
This mirrors the protocol contrast in Section~\ref{sec_prior_work}: hindsight-guided supervision helps when privileged context largely specifies what the model should produce, but does not reliably install reasoning states the student's distribution does not already support \citep{arxiv_2604_03128,DBLP:journals/corr/abs-2603-24472}.

Table~\ref{tab_marker_density} provides a lexical proxy for the less-exploratory side of this mechanism, measuring the density of epistemic markers of exploration and reflection per 1{,}000 response tokens for \{wait, hmm, perhaps, maybe, actually, alternatively, seems, might, likely, check\}, following an established marker set \citep{DBLP:journals/corr/abs-2603-24472}.
Both branches reduce these markers even after normalizing by response length, consistent with the view that the same compaction pressure removes hesitation, checking, and detours.
On correct rollouts, this can safely remove excess reasoning around an already-working solution; on incorrect rollouts, it removes the reconsiderations that a failed path still needs.

\begin{table}[H]
  \caption{Epistemic-marker density before and after OPSD on evaluation benchmarks, where density is the count of epistemic markers per 1{,}000 tokens. OPSD reduces marker density, consistent with the view that it removes redundant hesitation, over-verification, and detours from reasoning traces.}
  \label{tab_marker_density}
  \vspace{3pt}
  \centering
  \footnotesize
  \begin{tabular}{lccc}
    \toprule
    \textbf{Method} & \textbf{Baseline density} & \textbf{Post-OPSD density} & {\boldmath\textbf{$\Delta$Density (\%)}} \\
    \midrule
    \rowcolor{gray!15}\multicolumn{4}{l}{\textbf{Qwen3-8B}} \\
    Correct-only    & 10.61 & 8.15  & $-23.2$ \\
    Incorrect-only  & 10.61 & 9.41  & $-11.3$ \\
    \midrule
    \rowcolor{gray!15}\multicolumn{4}{l}{\textbf{AceReason-Nemotron-7B}} \\
    Correct-only    & 12.16 & 10.51 & $-13.5$ \\
    Incorrect-only  & 12.16 & 11.00 & $-9.5$  \\
    \bottomrule
  \end{tabular}
\end{table}

A remaining possibility is that \emph{Incorrect-only} fails because reverse-KL is mode-seeking: it keeps probability mass only where the self-teacher agrees with the student's current distribution, which may not force a failed trace toward teacher-supported alternatives \citep{DBLP:journals/corr/abs-2404-02657,DBLP:journals/corr/abs-2509-07430,DBLP:journals/corr/abs-2603-23871}.
Table~\ref{tab_main} therefore includes two divergence ablations, \emph{All-rollout JSD}, a symmetric divergence on every rollout, and \emph{Split-direction}, reverse-KL on correct rollouts and forward-KL on incorrect ones.
Both change the accuracy-length tradeoff without turning OPSD into a repair mechanism: \emph{All-rollout JSD} improves the AceReason-7B average over \emph{All-rollout} reverse-KL but still primarily compacts, and \emph{Split-direction} partially improves the Qwen3-8B tradeoff but does not transfer across benchmarks or models.
The limitation is not reverse-KL alone: the self-teacher still supplies a hindsight-conditioned signal on the student's own trajectory rather than a new sequence of reasoning states.

\section{Richer Context, Signal Reinjection, and Prolonged Training}\label{sec_ablations}

The outcome-filtered training results leave three narrower explanations to rule out.
First, OPSD may fail to repair because the self-teacher lacks sufficient privileged information.
Second, the teacher signal may be too front-loaded, weakening before it reaches the mid-rollout reasoning steps where repair is needed.
Third, repair may require many optimization steps before local token updates compound at the sequence level.
We test all three and find that they move variants along the same accuracy-length tradeoff rather than turning OPSD into a reliable repair mechanism.

\subsection{More Privileged Information Does Not Enable Repair}\label{sec_teacher_context}

One alternative explanation is that the \emph{Reflection from GT} privileged context is the binding constraint, either through the information it carries or through how it is structured.
If so, varying either axis should yield a separate accuracy regime distinct from the accuracy-length tradeoff observed in the main comparison.
Table~\ref{tab_teacher_context} reports six privileged-context variants on Qwen3-8B that span both axes, holding other configurations fixed.

\begin{table}[H]
  \caption{Privileged-context comparison on Qwen3-8B; each row varies the context shown to the self-teacher while keeping the loss fixed at \emph{All-rollout} reverse-KL. Richer or more direct privileged contexts mostly trade accuracy for compression rather than creating a separate repair regime. (M500=MATH500, A24=AIME24, A25=AIME25)}
  \label{tab_teacher_context}
  \vspace{3pt}
  \centering
  \footnotesize
  \setlength{\tabcolsep}{3pt}
  \begin{tabular}{lccccccccc}
    \toprule
    & \multicolumn{4}{c}{\textbf{Accuracy}} & \multicolumn{4}{c}{\textbf{Length}} & \\
    \cmidrule(lr){2-5} \cmidrule(lr){6-9}
    \textbf{Method} & \textbf{M500} & \textbf{A24} & \textbf{A25} & {\boldmath\textbf{Avg.\ $\Delta$\ (pp)}} & \textbf{M500} & \textbf{A24} & \textbf{A25} & {\boldmath\textbf{Avg.\ $\Delta$\ (\%)}} & \textbf{Loss} \\
    \midrule
    Baseline                  & $78.1$ & $70.8$ & $62.9$ & ---    & $4{,}849$ & $14{,}342$ & $17{,}031$ & ---     & ---      \\
    Conciseness instruction   & $79.3$ & $65.8$ & $59.2$ & $-2.5$ & $3{,}164$ & $11{,}620$ & $14{,}507$ & $-22.8$ & $0.0028$ \\
    Two-line Do/Don't         & $77.4$ & $73.8$ & $62.1$ & $+0.5$ & $4{,}582$ & $13{,}092$ & $16{,}181$ & $-6.4$  & $0.0149$ \\
    Four-line reflection      & $77.8$ & $73.8$ & $66.2$ & $+2.0$ & $4{,}270$ & $12{,}567$ & $15{,}833$ & $-10.4$ & $0.0217$ \\
    Reflection from GT        & $77.9$ & $69.6$ & $60.8$ & $-1.2$ & $3{,}419$ & $10{,}878$ & $12{,}499$ & $-26.8$ & $0.0401$ \\
    Reflection from demo      & $75.0$ & $72.5$ & $57.1$ & $-2.4$ & $3{,}502$ & $10{,}573$ & $13{,}177$ & $-25.6$ & $0.0414$ \\
    Re-solve from demo        & $79.5$ & $61.7$ & $50.8$ & $-6.6$ & $2{,}189$ & $8{,}466$  & $9{,}827$  & $-46.1$ & $0.0487$ \\
    \bottomrule
  \end{tabular}
\end{table}

In both \emph{Reflection from GT} and \emph{Reflection from demo}, the student attempts the problem, receives the bare final answer or a detailed worked solution from a more competent reasoning model, and writes a reflection.
Compression is essentially matched ($-26.8\%$ vs.\ $-25.6\%$ length) and accuracy is close ($-1.2$ vs.\ $-2.4$ pp).
Adding the worked solution does not open a separate accuracy regime despite the much richer reasoning content.
Information richness alone is therefore not the binding constraint.

A separate question is whether the reflection wrapper itself shapes the OPSD signal.
\emph{Re-solve from demo} hands the student the same expert demonstration but asks for a fresh solution, so the teacher-side context becomes a worked solution rather than a reflection.
The variant compresses much harder ($-46.1\%$ length) and loses more accuracy ($-6.6$ pp), moving aggressively along the same tradeoff rather than opening a separate repair regime.

\emph{Two-line Do/Don't} and \emph{Four-line reflection} move in the opposite direction: shorter structured reflections on the same expert demonstration, with a soft length constraint.
Their compression is correspondingly weaker ($-6.4\%$ and $-10.4\%$ length) and accuracy holds ($+0.5$ and $+2.0$ pp), at the gentler end of the same tradeoff.

The OPSD loss, the per-token gap that defines the advantage in Equation~\ref{eq_token_advantage}, tracks the structure of the privileged context rather than its information or its outcomes: \emph{Reflection from GT} and \emph{Reflection from demo} share nearly the same loss despite different information, while \emph{Conciseness instruction} and \emph{Reflection from demo} reach comparable outcomes from very different losses.
A larger gap means a stronger pull toward the teacher, not a more useful one.

Richer privileged contexts can make the self-teacher more hindsight-biased and less exploratory \citep{DBLP:journals/corr/abs-2603-24472}; distilled onto the student, this often strengthens compaction rather than repair.
Additional privileged information therefore moves variants along the same accuracy-length tradeoff, instead of supplying the missing repair signal.

\subsection{Signal Reinjection Cannot Repair Incorrect Rollouts}\label{sec_reinjection}

A second alternative explanation concerns where the teacher signal lands.
Because the privileged context appears only in the teacher's prompt prefix, the per-token reverse-KL signal may be strongest early and attenuate as the teacher's next-token distribution becomes dominated by the student's more recent tokens.
If so, OPSD may fail to repair because the teacher signal does not reach the mid-rollout reasoning steps where wrong trajectories need to change.

\begin{figure}[H]
  \makebox[\linewidth][c]{%
    \begin{minipage}{1.0\linewidth}
      \centering
      \includegraphics[width=1.0\linewidth]{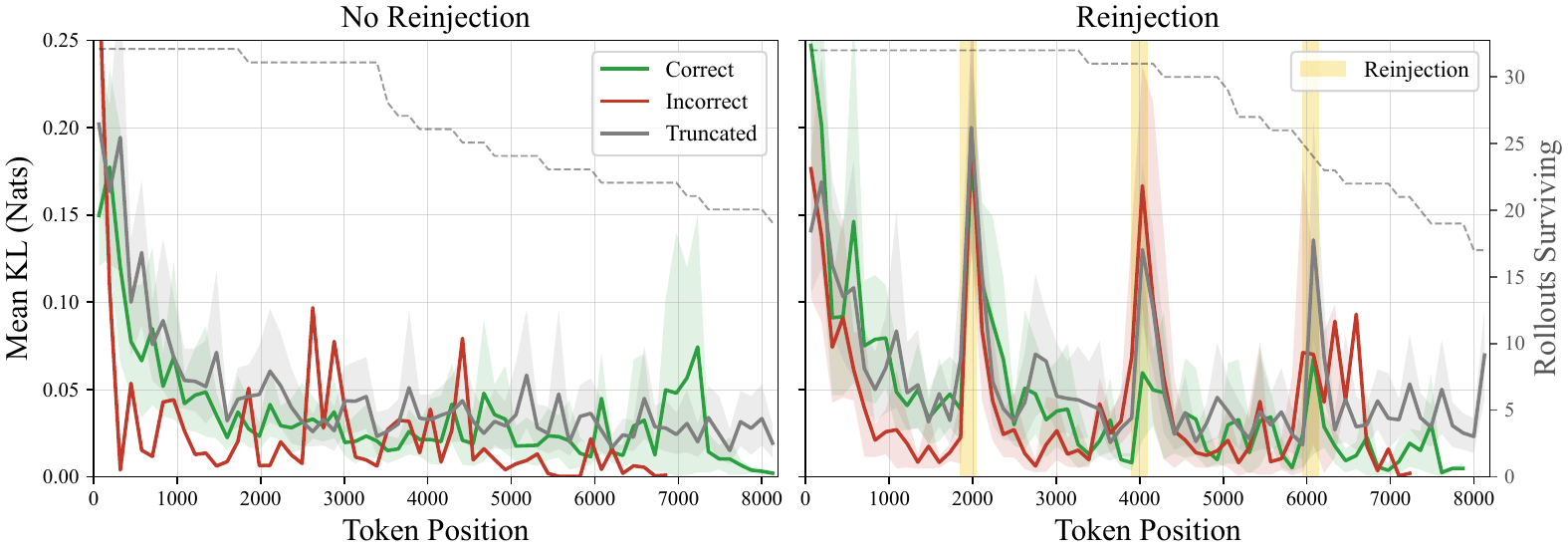}
      \caption{Per-token reverse-KL between student and teacher across rollout positions. The no-reinjection curves confirm that the teacher signal is front-loaded, and reinjection curves confirm the interventions land as intended. Curves are split by rollout outcome (\emph{Correct}, \emph{Incorrect}, \emph{Truncated}), with the dashed line tracking how many rollouts in each group survive to each position bin. Reshaping where the signal lands strengthens compression but does not reveal a failed-rollout repair channel.}
      \label{fig_reinjection}
    \end{minipage}
  }
\end{figure}

The intervention reinjects the privileged context into the teacher's input every 2{,}048 tokens within the student's reasoning trace, framed as a natural in-thinking recall (``Actually, I recall: \{privileged context\}. Continuing from this.''), while leaving the student's input unchanged.
We reinject \emph{Reflection from GT}.
Figure~\ref{fig_reinjection} confirms both the diagnosis and the intervention: without reinjection the teacher-student KL is front-loaded and decays with position, and with it the KL rises at the intended mid-trace locations.

Table~\ref{tab_reinjection} shows that reinjection changes the tradeoff in the direction of stronger compression: average length reduction increases from $-29.2\%$ without reinjection to $-33.1\%$ under \emph{Correct-only} reinjection.
However, the added mid-trace signal does not restore accuracy; the corresponding average accuracy change is $-2.6$ pp, and \emph{All-rollout} reinjection falls to $-4.2$ pp.
Outcome-filtered training still explains the larger qualitative difference: \emph{Correct-only} remains safer than \emph{All-rollout} under the same reinjection schedule, but neither becomes a failed-trajectory repair mechanism.
Thus the failure to repair does not appear to stem primarily from the teacher signal appearing only at the beginning of the rollout.

\begin{table}[H]
  \caption{Effect of privileged-context reinjection during OPSD training on Qwen3-8B. Reinjection strengthens compression but does not restore accuracy.}
  \label{tab_reinjection}
  \vspace{3pt}
  \centering
  \footnotesize
  \setlength{\tabcolsep}{2pt}
  \begin{tabular}{lcccccccc}
    \toprule
    & \multicolumn{4}{c}{\textbf{Accuracy}} & \multicolumn{4}{c}{\textbf{Length}} \\
    \cmidrule(lr){2-5} \cmidrule(lr){6-9}
    \textbf{Method} & \textbf{MATH500} & \textbf{AIME24} & \textbf{AIME25} & {\boldmath\textbf{Avg.\ $\Delta$\ (pp)}} & \textbf{MATH500} & \textbf{AIME24} & \textbf{AIME25} & {\boldmath\textbf{Avg.\ $\Delta$\ (\%)}} \\
    \midrule
    Baseline      & $78.1$ & $70.8$ & $62.9$ & ---    & $4{,}849$ & $14{,}342$ & $17{,}031$ & ---    \\
    \midrule
    \rowcolor{gray!15}\multicolumn{9}{l}{\textbf{No reinjection}} \\
    All-rollout   & $77.9$ & $69.6$ & $60.8$ & $-1.2$ & $3{,}419$ & $10{,}878$ & $12{,}499$ & $-26.8$ \\
    Correct-only  & $78.4$ & $72.1$ & $58.3$ & $-1.0$ & $3{,}208$ & $10{,}365$ & $12{,}587$ & $-29.2$ \\
    \midrule
    \rowcolor{gray!15}\multicolumn{9}{l}{\textbf{Reinjection every 2{,}048 tokens}} \\
    All-rollout   & $78.2$ & $68.8$ & $52.1$ & $-4.2$ & $3{,}454$ & $9{,}371$  & $11{,}243$ & $-32.5$ \\
    Correct-only  & $77.0$ & $71.2$ & $55.8$ & $-2.6$ & $3{,}103$ & $9{,}620$  & $11{,}829$ & $-33.1$ \\
    \bottomrule
  \end{tabular}
\end{table}

\subsection{Longer Training Adds No Repair, and Compaction Comes Cheap}\label{sec_longer_training}

A third alternative explanation is that repair needs many optimization steps before local token updates compound at the sequence level.
OPSD updates the student one token at a time toward a context-conditioned teacher, and several hundred such steps could in principle build up repair that 100 cannot.
We therefore extend the strongest 100-step accuracy variant, \emph{Split-direction}, to 500 steps.

\begin{wrapfigure}{r}{0.5\linewidth}
  \centering
  \vspace{-1.5\baselineskip}
  \includegraphics[width=\linewidth]{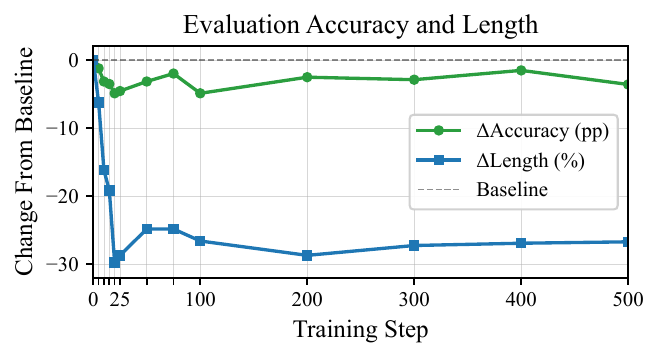}
  \vspace{-1.5\baselineskip}
  \caption{Extended OPSD training trajectory on Qwen3-8B, evaluated up to 500 steps. Length compression appears within the first 25 steps and persists, while accuracy oscillates below baseline.}
  \label{fig_500_steps}
  \vspace{-1.0\baselineskip}
\end{wrapfigure}

Figure~\ref{fig_500_steps} shows that extra OPSD budget preserves the length effect but does not accumulate a durable accuracy gain.
Length deltas drop to roughly $-25\%$ to $-30\%$ within the first 25 steps and stay in that band through step 500, while accuracy oscillates within a few pp below baseline ($-2.3$ pp at step 25 and again at step 500) without a durable upward trend.
This early saturation matches prior evidence that the useful effect of OPSD is concentrated early in training \citep{arxiv_2604_03128}.
Additional training therefore extends the same compaction regime for hundreds of further steps rather than converting OPSD into a delayed repair mechanism.

The early saturation is the practical appeal of OPSD as a compaction method: our 25-step run uses 32 rollouts per batch at an 8{,}192-token budget and already reaches the $-25\%$ to $-30\%$ compression regime.
Length-penalized RL recipes generate roughly $80\times$ to $2{,}000\times$ more rollouts than this budget \citep{arxiv_2601_21418, arxiv_2503_04697, arxiv_2504_01296}.
Short-trace RFT generates about $20\times$ to $200\times$ more rollouts in corpus construction before any fine-tuning \citep{DBLP:conf/emnlp/XiaLWLL25, arxiv_2509_14093, arxiv_2602_01982}.
The only directly comparable thinking-enabled work uses roughly $40\times$ more rollouts for ${-}17\%$ length reduction, less compression than our ${-}29\%$ \citep{arxiv_2602_01982}.
Table~\ref{tab_budget} tests both families at matched rollout budgets on the same prompts, and neither comes close: short-trace RFT barely shortens responses, and length-penalized GRPO improves accuracy but compresses a quarter as much despite a favorable eight rollouts per prompt and 12{,}288-token cap.

\begin{table}[H]
  \caption{Compression baselines on Qwen3-8B, averaged over MATH500, AIME24, and AIME25. \emph{Short-trace RFT} doubles the rollouts per step, keeps the verified-correct ones, and trains on the shorter half by length; \emph{Length-penalized GRPO} adds a DAPO-style soft-overlong penalty to the correctness reward. \emph{Rollouts} counts training generations. Neither baseline approaches OPSD's compression.}
  \label{tab_budget}
  \vspace{3pt}
  \centering
  \footnotesize
  \setlength{\tabcolsep}{4pt}
  \begin{tabular}{lcccccc}
    \toprule
    & & & \multicolumn{2}{c}{\textbf{Accuracy}} & \multicolumn{2}{c}{\textbf{Length}} \\
    \cmidrule(lr){4-5} \cmidrule(lr){6-7}
    \textbf{Method} & \textbf{Rollouts} & \textbf{Steps} & \textbf{Avg.} & {\boldmath\textbf{Avg.\ $\Delta$\ (pp)}} & \textbf{Avg.} & {\boldmath\textbf{Avg.\ $\Delta$\ (\%)}} \\
    \midrule
    Baseline              & ---       & ---   & $70.6$ & ---    & $12{,}074$ & ---     \\
    Correct-only OPSD     & $3{,}200$ & $100$ & $69.6$ & $-1.0$ & $8{,}720$  & $-29.2$ \\
    Short-trace RFT       & $3{,}200$ & $50$  & $70.1$ & $-0.5$ & $11{,}757$ & $-2.6$  \\
    Length-penalized GRPO & $3{,}200$ & $25$  & $74.1$ & $+3.5$ & $11{,}244$ & $-7.1$  \\
    \bottomrule
  \end{tabular}
\end{table}

\section{Conclusion}\label{sec_conclusion}

We asked whether OPSD's hindsight-guided signal, in thinking-enabled mathematical reasoning, ever repairs a failed trajectory or only compacts a viable one.
Across three models, training only on correct rollouts shortened responses while largely preserving accuracy, whereas training only on incorrect rollouts degraded it.
The gap held under a same-prompt control for difficulty, and neither branch raised the pass@$k$ ceiling.
Changing the divergence, enriching or reinjecting the privileged context, and training longer only moved OPSD along the same accuracy-length tradeoff.
Both branches also suppressed the reconsiderations that viable traces can spare but failed ones need, supporting our hypothesis that hindsight reveals which steps were unnecessary more readily than which would have repaired the solution.
Repairing failed reasoning calls for signals that discover new trajectories, such as outcome-based exploration, while OPSD cheaply compacts reasoning that is viable.
Hindsight compacts, but it does not repair.






\bibliography{references}
\bibliographystyle{iclr2027_conference}


\appendix

\clearpage
\section{Prompt-Matched Lockstep Training}\label{app_lockstep}

Rollout correctness is entangled with prompt difficulty, so the \emph{Correct-only} and \emph{Incorrect-only} branches in Table~\ref{tab_main} also train on different prompt distributions.
To remove the prompt-level split, we train the two branches in lockstep.
At every step both models draw the same candidate prompts and sample four rollouts per prompt; a prompt enters the step's batch only if the \emph{Correct-only} model produced at least one correct rollout and the \emph{Incorrect-only} model produced at least one incorrect rollout.
Each branch then trains on one qualifying rollout per admitted prompt with its own reverse-KL update, using \emph{Reflection from GT} as the privileged context.
Each step uses 16 active prompts rather than the 32 of Section~\ref{sec_setup}; other settings are unchanged.
Over 100 steps, the two branches queried 4{,}928 candidate prompts (39{,}424 rollout attempts) to admit 1{,}600 prompts each, a joint yield of 32.5\%, with 9.05M KL-active tokens on the \emph{Correct-only} branch and 9.99M on the \emph{Incorrect-only} branch.

\begin{table}[H]
  \caption{Prompt-matched lockstep training on Qwen3-8B. The unmatched rows repeat Table~\ref{tab_main}; the prompt-matched rows train the \emph{Correct-only} and \emph{Incorrect-only} branches on the same prompts at every step, admitting a prompt only if the \emph{Correct-only} model produces at least one correct rollout and the \emph{Incorrect-only} model at least one incorrect rollout within four attempts. Matching prompts preserves the ordering and most of the gap.}
  \label{tab_lockstep}
  \vspace{3pt}
  \centering
  \footnotesize
  \setlength{\tabcolsep}{2pt}
  \begin{tabular}{lcccccccc}
    \toprule
    & \multicolumn{4}{c}{\textbf{Accuracy}} & \multicolumn{4}{c}{\textbf{Length}} \\
    \cmidrule(lr){2-5} \cmidrule(lr){6-9}
    \textbf{Method} & \textbf{MATH500} & \textbf{AIME24} & \textbf{AIME25} & {\boldmath\textbf{Avg.\ $\Delta$\ (pp)}} & \textbf{MATH500} & \textbf{AIME24} & \textbf{AIME25} & {\boldmath\textbf{Avg.\ $\Delta$\ (\%)}} \\
    \midrule
    Baseline        & $78.1$ & $70.8$ & $62.9$ & ---     & $4{,}849$ & $14{,}342$ & $17{,}031$ & ---      \\
    \midrule
    \rowcolor{gray!15}\multicolumn{9}{l}{\textbf{Unmatched prompts (Table~\ref{tab_main})}} \\
    Correct-only    & $78.4$ & $72.1$ & $58.3$ & $-1.03$ & $3{,}208$ & $10{,}365$ & $12{,}587$ & $-29.22$ \\
    Incorrect-only  & $74.1$ & $68.3$ & $49.6$ & $-6.62$ & $3{,}957$ & $10{,}986$ & $13{,}723$ & $-20.41$ \\
    \midrule
    \rowcolor{gray!15}\multicolumn{9}{l}{\textbf{Prompt-matched lockstep}} \\
    Correct-only    & $78.1$ & $75.4$ & $59.6$ & $+0.42$ & $3{,}515$ & $10{,}965$ & $13{,}826$ & $-23.29$ \\
    Incorrect-only  & $72.4$ & $69.6$ & $59.6$ & $-3.44$ & $3{,}872$ & $11{,}782$ & $14{,}368$ & $-17.88$ \\
    \bottomrule
  \end{tabular}
\end{table}

Table~\ref{tab_lockstep} reports the result.
Matching prompts leaves the ordering intact and narrows the accuracy gap from 5.6 pp to 3.9 pp, while both branches compress somewhat less than in the unmatched setting.
The admitted prompts are those the model can solve within four attempts but does not solve reliably, so the remaining gap reflects the viability of the rollouts entering the KL objective rather than the difficulty of the prompts.

\clearpage
\section{Pass@$k$ Across Six Benchmarks}\label{app_passk}

A repair mechanism should raise the capability ceiling by making previously unsolvable questions solvable, whereas compaction adds no new solves.
We therefore compute pass@$k$ for $k\in\{1,2,4,8\}$ from the same eight samples per question used for average@8, per benchmark, and macro-average over six benchmarks (Table~\ref{tab_passk}).
\emph{Incorrect-only} is below the base at every $k$ on both models.
\emph{Correct-only} tracks the base on Qwen3-8B ($-0.58$ at pass@8); on AceReason-7B it raises pass@1 by 3.22 pp but falls below the base for $k\geq4$ ($-4.02$ at pass@8), which is what concentrating probability mass on already-viable paths looks like.
No branch raises the ceiling.

\begin{table}[H]
  \caption{pass@$k$ after 100 steps of outcome-filtered OPSD, estimated from eight samples per question, computed per benchmark, and macro-averaged over MATH500, AIME24, AIME25, AIME26, HMMT26, and OlympiadBench. No branch raises pass@8 above the base on either model.}
  \label{tab_passk}
  \vspace{3pt}
  \centering
  \footnotesize
  \begin{tabular}{lcccc}
    \toprule
    \textbf{Method} & \textbf{pass@1} & \textbf{pass@2} & \textbf{pass@4} & \textbf{pass@8} \\
    \midrule
    \rowcolor{gray!15}\multicolumn{5}{l}{\textbf{Qwen3-8B}} \\
    Baseline       & $66.58$ & $73.35$ & $77.36$ & $80.07$ \\
    Correct-only   & $65.38$ & $72.39$ & $76.43$ & $79.49$ \\
    Incorrect-only & $58.89$ & $67.16$ & $71.98$ & $74.92$ \\
    \midrule
    \rowcolor{gray!15}\multicolumn{5}{l}{\textbf{AceReason-Nemotron-7B}} \\
    Baseline       & $47.14$ & $58.63$ & $68.39$ & $76.74$ \\
    Correct-only   & $50.36$ & $59.53$ & $66.75$ & $72.72$ \\
    Incorrect-only & $41.83$ & $53.10$ & $63.01$ & $70.73$ \\
    \bottomrule
  \end{tabular}
\end{table}

\section{Question-Level Correctness Shift}\label{app_correctness_shift}

To check whether the \emph{Incorrect-only} degradation is merely an aggregate averaging artifact, we cross-tabulate pre- and post-OPSD question correctness on the combined evaluation set of MATH500, AIME24, and AIME25 (560 questions).
With base correctness $b=k_0/8$ and post-OPSD correctness $a=k/8$ for a question, its repaired mass is $(1-b)\,a$ and its damaged mass is $b\,(1-a)$; each cell reports the expected number of questions making that transition, and \emph{Net} is repaired minus damaged.
Table~\ref{tab_correctness_shift} shows the outcome-filtered branches: \emph{Incorrect-only} has strongly negative \emph{Net} in both main models, so the degradation reflects question-level damage.

\begin{table}[H]
  \caption{Question-level correctness shift after outcome-filtered OPSD on Qwen3-8B and AceReason-Nemotron-7B (MATH500, AIME24, and AIME25 combined; 560 questions). Cells are expected question counts from average@8 probabilities, with the share of the 560 questions in parentheses, and \emph{Net} is repaired minus damaged. \emph{Incorrect-only} has strongly negative \emph{Net} in both models, so its accuracy loss reflects question-level damage rather than only an aggregate averaging artifact.}
  \label{tab_correctness_shift}
  \vspace{3pt}
  \centering
  \footnotesize
  \setlength{\tabcolsep}{6pt}
  \begin{tabular}{lccccc}
    \toprule
    \textbf{Method} & \begin{tabular}[c]{@{}c@{}}\textbf{Repaired}\\\textbf{(I$\to$C)}\end{tabular} & \begin{tabular}[c]{@{}c@{}}\textbf{Damaged}\\\textbf{(C$\to$I)}\end{tabular} & \begin{tabular}[c]{@{}c@{}}\textbf{Stayed}\\\textbf{incorrect}\end{tabular} & \begin{tabular}[c]{@{}c@{}}\textbf{Stayed}\\\textbf{correct}\end{tabular} & \textbf{Net} \\
    \midrule
    \rowcolor{gray!15}\multicolumn{6}{l}{\textbf{Qwen3-8B}} \\
    Correct-only   & 38.7 (6.9\%)  & 38.5 (6.9\%)   & 90.5 (16.2\%)  & 392.3 (70.1\%) & $+0.2$ ($+0.0\%$)   \\
    Incorrect-only & 23.9 (4.3\%)  & 48.8 (8.7\%)   & 105.3 (18.8\%) & 381.9 (68.2\%) & $-24.9$ ($-4.4\%$)  \\
    All-rollout    & 31.0 (5.5\%)  & 35.6 (6.4\%)   & 94.8 (16.9\%)  & 398.7 (71.2\%) & $-4.6$ ($-0.8\%$)   \\
    \midrule
    \rowcolor{gray!15}\multicolumn{6}{l}{\textbf{AceReason-Nemotron-7B}} \\
    Correct-only   & 53.2 (9.5\%)  & 38.4 (6.9\%)   & 141.3 (25.2\%) & 327.1 (58.4\%) & $+14.8$ ($+2.6\%$)  \\
    Incorrect-only & 36.4 (6.5\%)  & 113.9 (20.3\%) & 158.1 (28.2\%) & 251.6 (44.9\%) & $-77.5$ ($-13.8\%$) \\
    \bottomrule
  \end{tabular}
\end{table}

The \emph{All-rollout} row applies the same analysis with no outcome gating of the training set: \emph{All-rollout} OPSD on Qwen3-8B for 100 steps, followed by the same cross-tabulation.
Repair and damage flows are comparable in size and \emph{Net} is $-4.6$ questions, so training on all rollouts produces question-level churn in both directions but no net repair channel.
On the same question set, average@8 moves by $-0.8$ pp while mean response length falls by $28.5\%$.

\clearpage
\section{Extended Benchmarks}\label{app_benchmarks}

Table~\ref{tab_benchmarks} extends the outcome-filtered comparison to AIME26, HMMT26, and OlympiadBench, alongside the three benchmarks of Table~\ref{tab_main}, for Qwen3-8B and AceReason-7B after 100 steps.
The \emph{Correct-only} versus \emph{Incorrect-only} ordering holds on every added benchmark for both models, and \emph{Correct-only} compresses more on average.
Both branches lose accuracy on HMMT26 on both models.

\begin{table}[H]
  \caption{Outcome-filtered OPSD on six benchmarks after 100 steps. Accuracy is average@8 in percent, length is mean generated tokens, and \emph{Avg.\ $\Delta$} averages per-benchmark changes from baseline, in percentage points for accuracy and in percent for length. Baseline rows come from the six-benchmark evaluation run and differ slightly from Table~\ref{tab_main}, which uses a separate evaluation of the same checkpoints. \emph{Correct-only} preserves accuracy better than \emph{Incorrect-only} on both models and on each of the three added benchmarks, and compresses more on average.}
  \label{tab_benchmarks}
  \vspace{3pt}
  \centering
  \footnotesize
  \setlength{\tabcolsep}{3pt}
  \begin{tabular}{lccccccc}
    \toprule
    \textbf{Method} & \textbf{MATH500} & \textbf{AIME24} & \textbf{AIME25} & \textbf{AIME26} & \textbf{HMMT26} & \textbf{OlympiadBench} & {\boldmath\textbf{Avg.\ $\Delta$}} \\
    \midrule
    \rowcolor{gray!15}\multicolumn{8}{l}{\textbf{Qwen3-8B, accuracy (\%)}} \\
    Baseline       & $78.9$ & $71.3$ & $63.3$ & $61.3$ & $45.1$ & $79.7$ & ---      \\
    Correct-only   & $79.9$ & $75.0$ & $60.4$ & $60.4$ & $39.4$ & $77.2$ & $-1.21$  \\
    Incorrect-only & $75.0$ & $65.0$ & $49.6$ & $53.8$ & $38.3$ & $71.7$ & $-7.69$  \\
    \rowcolor{gray!15}\multicolumn{8}{l}{\textbf{Qwen3-8B, length (tokens)}} \\
    Baseline       & $4{,}853$ & $14{,}317$ & $17{,}074$ & $15{,}564$ & $18{,}399$ & $10{,}511$ & ---      \\
    Correct-only   & $3{,}199$ & $9{,}929$  & $13{,}146$ & $11{,}978$ & $13{,}885$ & $7{,}452$  & $-27.40$ \\
    Incorrect-only & $4{,}202$ & $11{,}775$ & $13{,}521$ & $12{,}419$ & $14{,}741$ & $9{,}374$  & $-17.15$ \\
    \midrule
    \rowcolor{gray!15}\multicolumn{8}{l}{\textbf{AceReason-Nemotron-7B, accuracy (\%)}} \\
    Baseline       & $67.8$ & $39.6$ & $35.8$ & $36.7$ & $36.7$ & $66.3$ & ---      \\
    Correct-only   & $71.1$ & $46.3$ & $37.5$ & $40.8$ & $33.3$ & $73.2$ & $+3.22$  \\
    Incorrect-only & $53.2$ & $31.7$ & $25.8$ & $35.8$ & $33.0$ & $71.5$ & $-5.31$  \\
    \rowcolor{gray!15}\multicolumn{8}{l}{\textbf{AceReason-Nemotron-7B, length (tokens)}} \\
    Baseline       & $3{,}330$ & $9{,}043$ & $10{,}232$ & $9{,}492$ & $10{,}086$ & $6{,}426$ & ---      \\
    Correct-only   & $2{,}455$ & $7{,}521$ & $8{,}519$  & $8{,}012$ & $8{,}706$  & $5{,}097$ & $-18.30$ \\
    Incorrect-only & $2{,}701$ & $7{,}783$ & $8{,}386$  & $8{,}057$ & $8{,}532$  & $5{,}341$ & $-16.38$ \\
    \bottomrule
  \end{tabular}
\end{table}

\clearpage
\section{DeepSeek-R1-Distill-Qwen-7B, the Pre-RLVR Base of AceReason-Nemotron-7B}\label{app_lineage}

To extend the contrast to a model without RL training, we also ran \emph{Correct-only} and \emph{Incorrect-only} OPSD on R1-Distill-7B, the pre-RLVR base of AceReason-7B (Table~\ref{tab_r1}).
\emph{Correct-only} remains the safer branch, while \emph{Incorrect-only} remains negative on average accuracy.

\begin{table}[H]
  \caption{Outcome-filtered OPSD on DeepSeek-R1-Distill-Qwen-7B, the pre-RLVR base of AceReason-Nemotron-7B, evaluated on MATH500, AIME24, and AIME25. The same branch ordering holds without RLVR, with \emph{Correct-only} safer and \emph{Incorrect-only} negative on average accuracy.}
  \label{tab_r1}
  \vspace{3pt}
  \centering
  \footnotesize
  \setlength{\tabcolsep}{2pt}
  \begin{tabular}{lcccccccc}
    \toprule
    & \multicolumn{4}{c}{\textbf{Accuracy}} & \multicolumn{4}{c}{\textbf{Length}} \\
    \cmidrule(lr){2-5} \cmidrule(lr){6-9}
    \textbf{Method} & \textbf{MATH500} & \textbf{AIME24} & \textbf{AIME25} & {\boldmath\textbf{Avg.\ $\Delta$\ (pp)}} & \textbf{MATH500} & \textbf{AIME24} & \textbf{AIME25} & {\boldmath\textbf{Avg.\ $\Delta$\ (\%)}} \\
    \midrule
    Baseline           & $65.5$ & $38.8$ & $26.2$ & ---    & $2{,}886$ & $10{,}268$ & $11{,}724$ & ---    \\
    Correct-only    & $65.8$ & $40.4$ & $30.0$ & $+1.9$ & $2{,}256$ & $8{,}572$  & $9{,}112$  & $-20.2$ \\
    Incorrect-only  & $60.4$ & $32.9$ & $25.4$ & $-3.9$ & $2{,}429$ & $8{,}714$  & $9{,}578$  & $-16.4$ \\
    \bottomrule
  \end{tabular}
\end{table}

\section{Multi-Seed Robustness of the Correct-only versus Incorrect-only Contrast}\label{app_seeds}

As a complementary check, we reran \emph{Correct-only} and \emph{Incorrect-only} on both main models with two additional training seeds (Table~\ref{tab_multiseed}).
\emph{Correct-only} is better than \emph{Incorrect-only} in all matched model-seed comparisons, so the branch ordering is stable across seeds.

\begin{table}[H]
  \caption{Multi-seed robustness of the \emph{Correct-only} versus \emph{Incorrect-only} contrast, averaged over MATH500, AIME24, and AIME25; \emph{seed 42} is the run reported in Table~\ref{tab_main}, and \emph{seed 43} and \emph{seed 44} are independent reruns varying the training seed. \emph{Correct-only} is better than \emph{Incorrect-only} in every model-seed pair, so the branch ordering is stable across seeds.}
  \label{tab_multiseed}
  \vspace{3pt}
  \centering
  \footnotesize
  \setlength{\tabcolsep}{2pt}
  \begin{tabular}{lcccccccc}
    \toprule
    & \multicolumn{4}{c}{{\boldmath\textbf{$\Delta$Accuracy (pp)}}} & \multicolumn{4}{c}{{\boldmath\textbf{$\Delta$Length (\%)}}} \\
    \cmidrule(lr){2-5} \cmidrule(lr){6-9}
    \textbf{Method} & \textbf{seed 42} & \textbf{seed 43} & \textbf{seed 44} & {\boldmath\textbf{Mean $\pm$ std}} & \textbf{seed 42} & \textbf{seed 43} & \textbf{seed 44} & {\boldmath\textbf{Mean $\pm$ std}} \\
    \midrule
    \rowcolor{gray!15}\multicolumn{9}{l}{\textbf{Qwen3-8B}} \\
    Correct-only    & $-1.03$  & $-3.40$  & $-0.39$   & $-1.61 \pm 1.58$  & $-29.22$ & $-29.74$ & $-30.05$ & $-29.67 \pm 0.42$ \\
    Incorrect-only  & $-6.62$  & $-6.89$  & $-10.33$  & $-7.94 \pm 2.07$  & $-20.41$ & $-18.52$ & $-22.12$ & $-20.35 \pm 1.80$ \\
    \midrule
    \rowcolor{gray!15}\multicolumn{9}{l}{\textbf{AceReason-Nemotron-7B}} \\
    Correct-only    & $+3.46$  & $+2.59$  & $+4.51$   & $+3.52 \pm 0.96$  & $-17.61$ & $-19.64$ & $-16.65$ & $-17.96 \pm 1.53$ \\
    Incorrect-only  & $-10.00$ & $-12.13$ & $-16.53$  & $-12.88 \pm 3.33$ & $-17.35$ & $-16.09$ & $-17.92$ & $-17.12 \pm 0.94$ \\
    \bottomrule
  \end{tabular}
\end{table}

\clearpage
\section{Privileged Context Examples}\label{app_teacher_context_examples}

This appendix shows the privileged content the OPSD self-teacher was conditioned on for each of the six privileged-context variants in Table~\ref{tab_teacher_context}. For \emph{Conciseness instruction}, the privileged content is a fixed prompt prepended to the teacher's input. For the other five variants, we show the math problem the student was given alongside the resulting reflection (or, in the case of \emph{Re-solve from demo}, the resulting worked solution) that the self-teacher then read.

\subsection{Conciseness instruction}

Following prior work~\citep{arxiv_2603_05433}, we prepend the instruction below to the teacher's input for every problem.

\begin{lstlisting}[style=promptbox]
Solve the following math problem concisely and correctly. Be direct -- avoid unnecessary elaboration, redundant steps, or restating the problem. Focus only on the key reasoning steps needed to reach the answer.
\end{lstlisting}

\subsection{Two-line Do/Don't}

\promptlabel{Problem.}
\begin{lstlisting}[style=promptbox]
A laser is placed at the point $(3,5)$. The laser beam travels in a straight line. Larry wants the beam to hit and bounce off the $y$-axis, then hit and bounce off the $x$-axis, then hit the point $(7,5)$. What is the total distance $d$ the beam will travel along this path? If the answer is $a\sqrt{b}$, where $a$ and $b$ are integers and $b$ is not divisible by the square of any prime, what is the value of $a$?
\end{lstlisting}

\promptlabel{Reflection.}
\begin{lstlisting}[style=promptbox]
* DO: Reflect the final point across the axes in reverse order of the bounces and compute the straight-line distance between the original starting point and the doubly-reflected image.  
* DON'T: Attempt to model each bounce as a separate segment with angle calculations; this obscures the geometric relationship and introduces unnecessary complexity.
\end{lstlisting}

\subsection{Four-line reflection}

\promptlabel{Problem.}
\begin{lstlisting}[style=promptbox]
What is $3^{-1} + 3^{-2} \pmod{25}$? Express your answer as an integer from $0$ to $24$, inclusive.
\end{lstlisting}

\promptlabel{Reflection.}
\begin{lstlisting}[style=promptbox]
Key insight: The problem hinges on recognizing that negative exponents in modular arithmetic correspond to modular inverses, which can be calculated using the Extended Euclidean Algorithm or trial multiplication, provided the base and modulus are coprime.

Recognition pattern: The presence of negative exponents ($a^{-1}, a^{-2}$) and a modulus (25) signals the need to compute modular inverses. The structure of the expression suggests leveraging exponent rules (e.g., $a^{-2} = (a^{-1})^2$) to simplify the computation.

What you did right: You correctly identified that $3^{-2}$ can be expressed as the square of $3^{-1}$, reducing the problem to computing a single inverse and squaring it, rather than finding the inverse of 9 directly, which simplifies the process.

Efficient path: Identify the need for modular inverses, compute $a^{-1}$ via multiplication or algorithm, square it to get $a^{-2}$, sum the results, and reduce modulo $n$ using division or subtraction.
\end{lstlisting}

\subsection{Reflection from GT}

\promptlabel{Problem.}
\begin{lstlisting}[style=promptbox]
If the product $\left(4^5\right)\left(5^{13}\right)$ is written as an integer, how many digits are in this integer?
\end{lstlisting}

\promptlabel{Reflection.}
\begin{lstlisting}[style=promptbox]
Memo: Key Insights and Efficient Path for Solving the Problem  

**Key Insight:**  
The critical insight was recognizing that $4^5$ can be rewritten as $2^{10}$, allowing the expression $2^{10} \cdot 5^{13}$ to be simplified by grouping $2^{10}$ and $5^{10}$ into $10^{10}$. This reduces the problem to calculating $10^{10} \cdot 5^3 = 125 \cdot 10^{10}$, which is straightforward to evaluate.  

**Efficient Path:**  
1. **Rewrite bases:** Convert $4^5$ into $2^{10}$ to create a common factor with $5^{13}$.  
2. **Pair exponents:** Combine $2^{10} \cdot 5^{10} = 10^{10}$ and leave $5^3$ as a separate factor.  
3. **Simplify:** Multiply $10^{10} \cdot 125$ to get $125$ followed by ten zeros.  
4. **Count digits:** Recognize that $125$ (3 digits) plus ten zeros yields a total of **13 digits**.  

**Patterns to Recognize:**  
- **Pair 2s and 5s:** Whenever a product involves $2^n \cdot 5^m$, pairing $2^k \cdot 5^k = 10^k$ simplifies the problem.  
- **Logarithmic verification:** For larger exponents, using $\lfloor \log_{10} N \rfloor + 1$ ensures accuracy.  
- **Structure of $10^k$:** Numbers like $10^{10}$ are easy to visualize as $1$ followed by ten zeros, which aids in digit counting.  

By mastering these steps and patterns, you can efficiently solve similar problems involving exponent products and digit counts.  

Answer: $13$
\end{lstlisting}

\subsection{Reflection from demo}

\promptlabel{Problem.}
\begin{lstlisting}[style=promptbox]
At the grocery store, I bought $4$ different items. I brought $3$ identical bags, and handed them to the cashier. How many ways are there for the cashier to put the items I bought in the $3$ identical bags, assuming he might leave some of the bags empty?
\end{lstlisting}

\promptlabel{Reflection.}
\begin{lstlisting}[style=promptbox]
### Memo: Key Insights and Efficient Approach

---

### **Key Insight**

The problem involves **distributing distinct items into identical bags** with **allowed empty bags**, which is a classic case of **set partitioning**. The critical realization is that **bags are not distinct**, so the focus is on grouping items rather than assigning them to labeled containers. Therefore, **Stirling numbers of the second kind** $ S(n, k) $, which count the number of ways to partition a set of $ n $ elements into exactly $ k $ **non-empty** subsets, are the correct mathematical tool.

---

### **Most Efficient Path**

1. **Challenge**: Identical bags (no ordering) and the possibility of empty bags complicate the usual distribution methods (e.g., permutations or combinations with labeled containers).
2. **Solution Strategy**:
   - Recognize that even with empty bags allowed, the task reduces to counting all **set partitions** of the 4 items into 1, 2, or 3 subsets (since there are 3 bags).
   - Use **Stirling numbers of the second kind**, summing $ S(4, 1) + S(4, 2) + S(4, 3) $ to account for all groupings.

---

### **Structural Features to Recognize**

- **Identical containers**: This is a strong indicator that **bag identity does not matter**, and the focus should be on **grouping** (not labeling).
- **Distinct items**: The number of ways depends on how the items are **divided into subsets**.
- **Empty bags allowed**: This implies that the sum of the number of non-empty subsets must be <= the number of bags.

---

### **Why This Works**

The strategy of summing Stirling numbers directly addresses the core of the problem: **how many distinct ways to partition a set of size $ n $** into up to $ k $ subsets when the subsets are **indistinguishable**. This avoids overcounting due to repeated groupings (e.g., swapping bags doesn't yield a new arrangement), which is a common pitfall with labeled bags.

---

### Final Summary

By recognizing that the problem is a **set partitioning** task and applying the **Stirling numbers of the second kind**, the solution becomes straightforward and efficient. This insight avoids unnecessary overcomplication and directly leads to the correct total of **14 distinct ways** to distribute 4 items into 3 identical bags with empty bags allowed.

---

Answer: 14
\end{lstlisting}

\subsection{Re-solve from demo}

\promptlabel{Problem.}
\begin{lstlisting}[style=promptbox]
A rectangle that is inscribed in a larger rectangle (with one vertex on each side) is called unstuck if it is possible to rotate (however slightly) the smaller rectangle about its center within the confines of the larger. Of all the rectangles that can be inscribed unstuck in a 6 by 8 rectangle, the smallest perimeter has the form $\sqrt{N}\,$, for a positive integer $N\,$. Find $N\,$.
\end{lstlisting}

\promptlabel{Worked solution.}
\begin{lstlisting}[style=promptbox]
To find the smallest perimeter of an **unstuck** inscribed rectangle within a **6 by 8** rectangle, we analyze the geometric constraints imposed by the requirement that the smaller rectangle must be able to rotate slightly about its center.

---

### Key Observations:

- An **unstuck** rectangle must not be axis-aligned or degenerate. This means a small rotation is possible, so the rectangle must be rotated and not "locked" in place.
- If a rectangle is inscribed such that one vertex lies on each side of the larger rectangle, the coordinates of its vertices can be parameterized relative to the center of the larger rectangle.
- Using geometric reasoning and symmetry, we consider the case where a rectangle becomes **just unstuck**, i.e., at the threshold where it is not axis-aligned, but can still be rotated slightly.
- The most efficient and minimal perimeter configuration occurs at this threshold, where the side lengths are derived from the rotation angle and geometric constraints.

---

### Calculation:

Using symmetry and optimality conditions, we find that the minimal perimeter corresponds to a rectangle whose side lengths, derived from the center and rotation, lead to the perimeter expression:

$$
P = 8\sqrt{7}
$$

This simplifies to:

$$
P = \sqrt{448}
$$

---

### Final Answer:

$$
\boxed{448}
$$
\end{lstlisting}


\end{document}